\documentclass[letterpaper, 10 pt, conference]{ieeeconf}  
  \usepackage{pgfplots}
  \pgfplotsset{compat=newest}
  \usetikzlibrary{plotmarks}
  \usetikzlibrary{arrows.meta}
  \usepgfplotslibrary{patchplots}
  \usepackage{grffile}
  \usepackage{amsmath}
  \usepackage{blindtext}
  \usepackage[bottom]{footmisc}
  \usepackage{mathrsfs}
  \usepackage{lipsum}
 \usepackage{url}
 \usepackage{array}
\usepackage{booktabs}
\usepackage{float}
\usepackage{multirow}
\usepackage{cite}

\usepackage{svg}

\newlength\fwidth
 \usepackage[linesnumbered,ruled]{algorithm2e}

\usepackage{graphicx}
\usepackage{framed}
\usepackage{indentfirst}
\usepackage{multirow}
\usepackage{makecell}

\usepackage{amsmath, amssymb,amsthm}
\usepackage{color}
\usepackage{booktabs}
\usepackage{tabularx}
\usepackage{hyperref}
\usepackage[capitalise]{cleveref}
\usepackage{comment}

\usepackage{makecell}
\let\labelindent\relax
\usepackage{enumitem}

\setcellgapes{1pt}

\newtheoremstyle{exampstyle}
{2pt} 
{2pt} 
{\itshape} 
{} 
{\bfseries} 
{.} 
{.5em} 
{\thmname{#1}\thmnumber{#2}\thmnote{(#3)}} 
\theoremstyle{exampstyle}
\newtheorem{hypothesis}{H}

\usepackage{hyperref}

\newcommand{\eref}[1]{Eq.~\eqref{#1}}  
\newcommand{\figref}[1]{Fig.~\ref{#1}}  
\newcommand{\tabref}[1]{Table~\ref{#1}} 
\newcommand{\prg}[1]{\noindent\textbf{#1}} 

\IEEEoverridecommandlockouts                              

\usepackage[colorinlistoftodos]{todonotes}

\newcommand{\mx}[1]{\textcolor{black}{#1}}

\usepackage[font=footnotesize]{caption}
\usepackage{balance}
\usepackage{hyperref}
 \hypersetup{
     colorlinks=true,
     linkcolor=orange,
     filecolor=orange,
     citecolor=orange,      
     urlcolor=orange,
     }

\title{\LARGE \bf
Learning Human Objectives from Sequences of Physical Corrections
}

\author{Mengxi Li$^{1}$, Alper Canberk$^{1}$, Dylan P. Losey$^{2}$, Dorsa Sadigh$^{1}$
\thanks{$^{1}$Intelligent and Interactive Autonomous Systems Group (\href{https://iliad.stanford.edu/}{ILIAD}), Dept of Computer Science, Stanford University, Stanford, CA 94305. \newline
$^{2}$ Collaborative Robotics Lab (\href{https://collab.me.vt.edu/}{Collab}), Virginia Tech.
\newline
{(e-mail: mengxili@stanford.edu)}}%
}

\begin{document}

\maketitle
\thispagestyle{empty}
\pagestyle{empty}

\begin{abstract}

When personal, assistive, and interactive robots make mistakes, humans naturally and intuitively correct those mistakes through physical interaction. In simple situations, one correction is sufficient to convey what the human wants. But when humans are working with multiple robots or the robot is performing an intricate task often the human must make \textit{several} corrections to fix the robot's behavior. Prior research assumes each of these physical corrections are \textit{independent} events, and learns from them one-at-a-time. However, this misses out on crucial information: each of these interactions are \textit{interconnected}, and may only make sense if viewed together. Alternatively, other work reasons over the \textit{final} trajectory produced by all of the human's corrections. But this method must wait until the end of the task to learn from corrections, as opposed to inferring from the corrections in an online fashion. In this paper we formalize an approach for learning from sequences of physical corrections during the current task. To do this we introduce an auxiliary reward that captures the human's trade-off between making corrections which improve the robot's immediate reward and long-term performance. We evaluate the resulting algorithm in remote and in-person human-robot experiments, and compare to both \textit{independent} and \textit{final} baselines. Our results indicate that users are best able to convey their objective when the robot reasons over their sequence of corrections.

\end{abstract}

\section{Introduction}
\label{intro}

\begin{figure}[t]
    \vspace{0.5em}
	\begin{center}
		\includegraphics[width=\columnwidth]{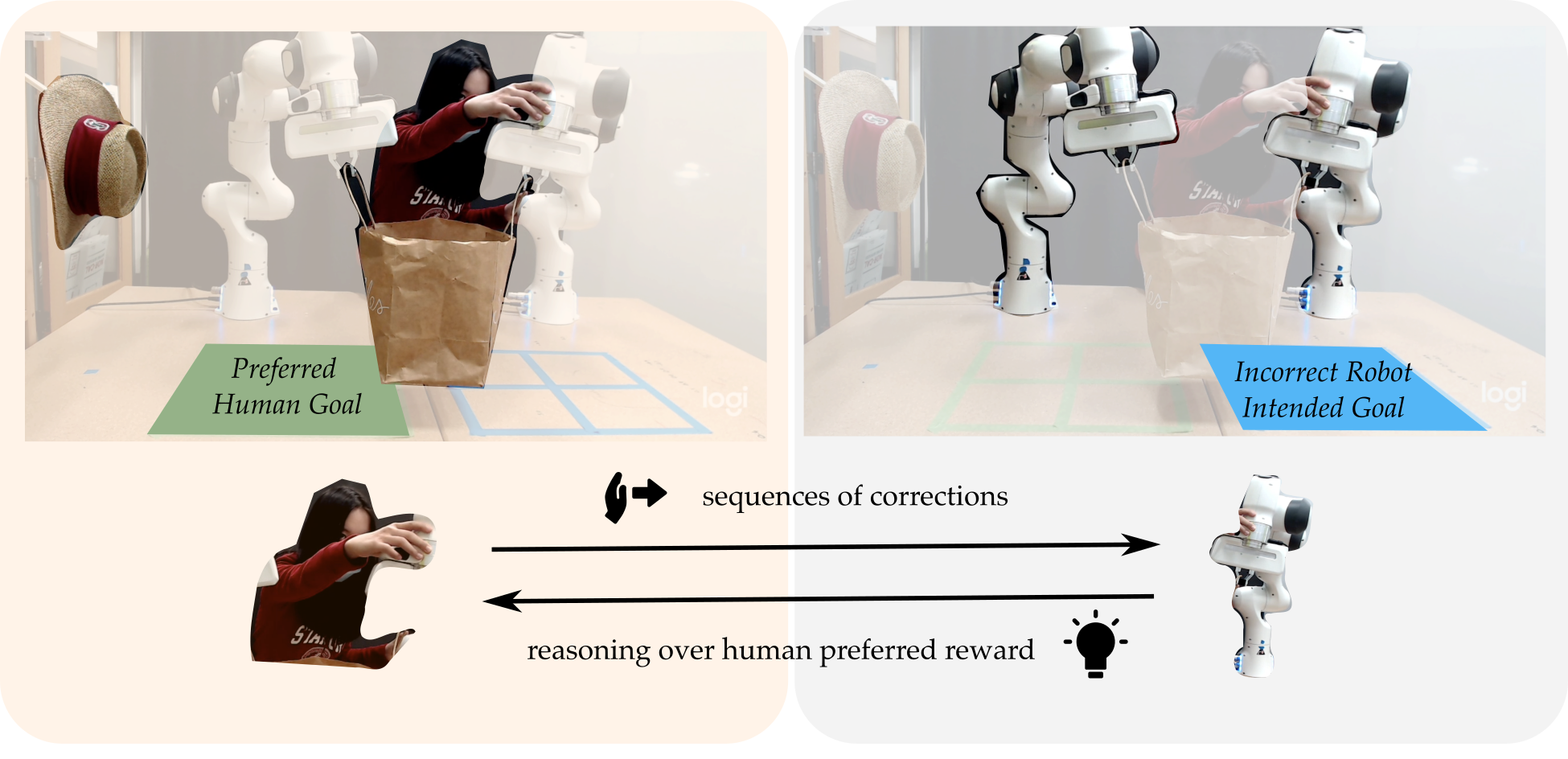}

		\caption{Learning from physical sequences. Two robot arms are carrying a grocery bag toward an undesirable wet region, blue region on the right. The human provides a sequence of physical corrections to guide the robot toward their preferred objective, i.e., placing the bag on the green region while also avoiding the obstacles on the left, and holding the bag upright without squeezing or stretching it. }

		\label{fig:front}
	\end{center}
	
	\vspace{-2em}
	
\end{figure}

Imagine that you just got back from the store, and a two-armed personal robot is helping you unpack a bag of groceries (see Fig.~\ref{fig:front}). You don't want this robot to bump the bag into any cabinets or the hat on the left, but you also don't want the robot to stretch and rip the bag, or squeeze it and crush your groceries. Out of the corner of your eye, you notice that the robot is heading towards a side of the table that is wet (the blue region in the figure). So you \textit{physically correct} it --- pushing, pulling, or twisting the robot's arms to guide it away from that region and move it toward the green region on the left. Importantly, you can only physically interact with one arm at a time, since it's difficult to pay attention to how to correct both arms simultaneously: in the process of guiding this arm away from the obstacle, you might inadvertently move both arms closer together, squeezing the bag. Teaching this robot about all your preferences requires more than just one physical correction. You must provide a \textit{sequence}: alternating between fixing the position of the arm closest to the obstacle, and adjusting the other arm so that the bag is held correctly.

State-of-the-art methods \textit{learn} from physical corrections by treating each interaction as an \textit{independent} event \cite{bajcsy2017learning, bajcsy2018learning, losey2018including, bobu2020quantifying}. These works assume that the human makes corrections based only on their objective, without considering the other corrections they have already made or are planning to provide. But if we view human corrections as isolated events, we can misinterpret what they convey: for instance, when the human moves one arm away from the hat and closer to the other arm, this robot will mistakenly learn that the human wants to squeeze the bag.

At the other end of the spectrum, robots can learn by looking at the \textit{final} trajectory collectively produced by all of the human's corrections \cite{akgun2012keyframe, argall2009survey, osa2018algorithmic, ratliff2006maximum, jain2015learning}. There are two issues with this: i) the robot does not learn or update its behavior until after the entire task is over, and ii) even the final trajectory may not capture what the human really wants. Returning to our example: because the user can only interact with one arm at a time, the final trajectory has some parts where the distance to the obstacle is right, and other sections where the bag is held correctly --- but the final trajectory fails to capture both throughout.

At their core, prior works miss out on part of the process that humans use to correct the robot's behavior. Not everything can be fixed at once, or even fixed perfectly:
\begin{center}
\vspace{-0.5em}
    \textit{Humans corrections are not independent events --- we often use multiple correlated interactions to correct the robot.}
\vspace{-0.5em}
\end{center}
We leverage this insight to learn the human's reward function online from sequences of physical corrections, \textit{without assuming} that human corrections are conditionally independent. Let's jump back to our example: a robot that reasons over the sequence recognizes that \textit{collectively} the human corrections keep the robot away from the obstacle while holding the bag, even though the corrections \textit{individually} fail to convey this objective, and can even be counter-productive.

Overall, we make the following contributions:

\noindent\textbf{Capturing Conditional Dependence.} We enable robots to learn from sequences of corrections by introducing an auxiliary reward function. This reward captures the human's trade-off between making corrections that increase the short-term reward (i.e., avoiding the obstacle) and reaching their long-term objective (i.e., carrying groceries).

\smallskip

\noindent\textbf{Learning from Sequences.} We introduce a tractable method to learn from a sequence of physical corrections by i) using the Laplace approximation to estimate the partition function and ii) solving a mixed-integer optimization problem.

\smallskip

\noindent\textbf{Conducting Online and In-Person User Studies.} Participants interacted with robots in both single- and multi-agent environments. We recorded user's corrections, and compared our approach to both \textit{independent} and \textit{final} baselines. Our proposed method outperforms both baselines, demonstrating the effectiveness of reasoning over sequences.

\section{Related Work}
\label{sec:related_work}

\noindent\textbf{Physical Human-Robot Interaction.} When humans and robots share a workspace, physical interaction is \textit{inevitable}. Prior work studies how robots can safely respond to physical interactions \cite{haddadin2016physical, de2008atlas, ikemoto2012physical}. This includes impedance control \cite{hogan1985impedance} and other reactive strategies \cite{haddadin2008collision}. Most relevant to our setting is \textit{shared control} \cite{li2015continuous, losey2018review, abbink2018topology, mortl2012role}, where the human and robot arbitrate between leader and follower roles. Although shared control enables the user to temporarily correct the robot's motion, it does not alter the robot's long term behavior: the robot does not \textit{learn} from physical corrections.

\smallskip

\noindent\textbf{Learning from Corrections (Online).} Recent research recognizes that physical human corrections are often \textit{intentional}, and therefore informative \cite{bajcsy2017learning, bajcsy2018learning, losey2018including, bobu2020quantifying}. These works learn about the human's underlying objective in real-time by comparing the current correction to the robot's previous behavior. Importantly, each correction is treated as an \textit{independent} event. Outside of physical human-robot interaction, shared autonomy follows a similar learning scheme --- the robot uses human inputs to update its understanding of the human's goal online, but does not reason about the connections between multiple interactions \cite{javdani2018shared, dragan2013policy, jain2019probabilistic,jeon2020shared}. We build upon these prior works by learning from sequences of corrections.

\smallskip

\noindent\textbf{Learning from Corrections (Offline).} By contrast, other works learn from the \textit{final} trajectory produced after all the corrections are complete. This research is closely related to learning from demonstrations \cite{argall2009survey}. For example, in \cite{akgun2012keyframe} the human corrects keyframes along the robot's trajectory, so that the next time the robot encounters the same task it moves through the corrected keyframes. Most similar to our setting are \cite{jain2015learning} and \cite{ratliff2006maximum}, where the robot iteratively updates its understanding of the human's objective \textit{after} the human corrects the robot's entire trajectory. Although these works take multiple corrections into account, they do so offline, and are not helpful during the current task.

\section{Formalizing Sequences of Physical Corrections}
\label{sec:formalism}

In this section we formalize a physical human-robot interaction setting where one or more robots are performing a task incorrectly. The human expert knows how these robots should behave, and physically corrects the robots to convey the true objective. But the human doesn't interact just once: the human may need to interact \textit{multiple times} in order to correct the robots. Our goal is for these robots to learn the human's objective from this sequence of physical corrections.

\subsection{Task Formulation} 
\label{sec:task_formulation}

We formulate our problem as a discrete-time Markov Decision Process (MDP) $\mathcal{M}=\left(\mathcal{S}, \mathcal{A}, \mathcal{T}, r, \gamma, \rho_{0}\right)$. Here $\mathcal{S} \subseteq \mathbb{R}^{n}$ is the \mx{robot} state space, $\mathcal{A} \subseteq \mathbb{R}^{m}$ is the robot action space, $\mathcal{T}(s, a)$ is the transition probability function, $r$ is the reward, $\gamma$ is the discount factor, and $\rho_{0}$ is the initial distribution.

\smallskip
\prg{Reward.}
Let the robot start from a state $s^0$ at time $t=0$. As the robot completes the task it follows a trajectory of states: $\xi = \{s^0, s^1, \dots, s^T\} \in \Xi$. The human has in mind a trajectory that they prefer for the robot to follow. Recall our motivating example --- here the human wants the robot arms to follow a trajectory that avoids the cabinets without squashing the bag. Similar to prior work \cite{ziebart2008maximum, abbeel2004apprenticeship, osa2018algorithmic, jeon2020reward}, we capture this objective through our reward function:
$R(\xi; \theta) = \theta \cdot \Phi(\xi)$.
Here $\Phi$ denotes the feature counts over trajectory $\xi$, and $\theta$ captures how important each feature is to the human. We let $\xi_R^t$ denote the robot's trajectory at timestep $t$, and we let $\theta^t$ denote the robot's current reward weights.

\smallskip
\prg{Suboptimal Initial Trajectory.} The system of one or more robots starts off with an initial reward function $R(\xi; \theta^0)$, and optimizes this reward function to produce its initial trajectory.
$$\xi_R^{0}=\arg \min _{\xi \in \Xi} \theta^{0} \cdot \Phi(\xi)$$
But this initial trajectory $\xi_R^0$ misses out on what the human really wants --- going back to our example, the robot does not realize that the blue region is wet and it needs to place the bag on the green region.
More formally, the robot's estimated reward function (which is parameterized by $\theta^0$) does not match the human's preferred reward function (parameterized by the true weights $\theta^*$).

\smallskip
\prg{Human Corrections.} The robot learns about the human's reward --- i.e., the true reward weights --- from physical corrections. Intuitively, these corrections are applied forces and torques which push, twist, and guide the robots. To formulate these interactions we must revise our problem definition: let $a_R$ be the robot's action and let $a_H$ be the human's \textit{correction}. In practice, both $a_R$ and $a_H$ could be applied joint torques \cite{bajcsy2017learning, bajcsy2018learning}. Now the overall system transitions based on both human and robot actions: $s^{t+1} = \mathcal{T}\left(s^t, a_{R}+a_{H}\right)$. We use $A_H = \{(t_i, a_H^i), i=1,\ldots, K\}$ to denote a \textit{sequence} of $K$ ordered human corrections $a_H^i$ at time step $t_i$, where $i$ keeps track of order of the corrections.
Our goal is to learn the human's true reward weights from the sequence of corrections $A_H$.



\subsection{Physical Corrections as Observations}
\label{sec:observation}

When robots are performing a task suboptimally, the human expert intervenes to correct those robots towards the right behavior. Going back to our example from Fig.~\ref{fig:front}. The user sees that the robot is making a mistake (moving towards the wet blue region), and physically intervenes. In the process of fixing this first issue, the human is forced to create another problem: by moving the first robot arm away from the blue region, they also move both arms closer together, and start to squash the bag. We note two important characteristics of these corrections: i) each human correction is intentional, and conveys information about the human's objective, but ii) the corrections viewed together may provide more information than isolating each interaction.

\begin{figure*}[t]
	\begin{center}
		\includegraphics[width=1.8\columnwidth]{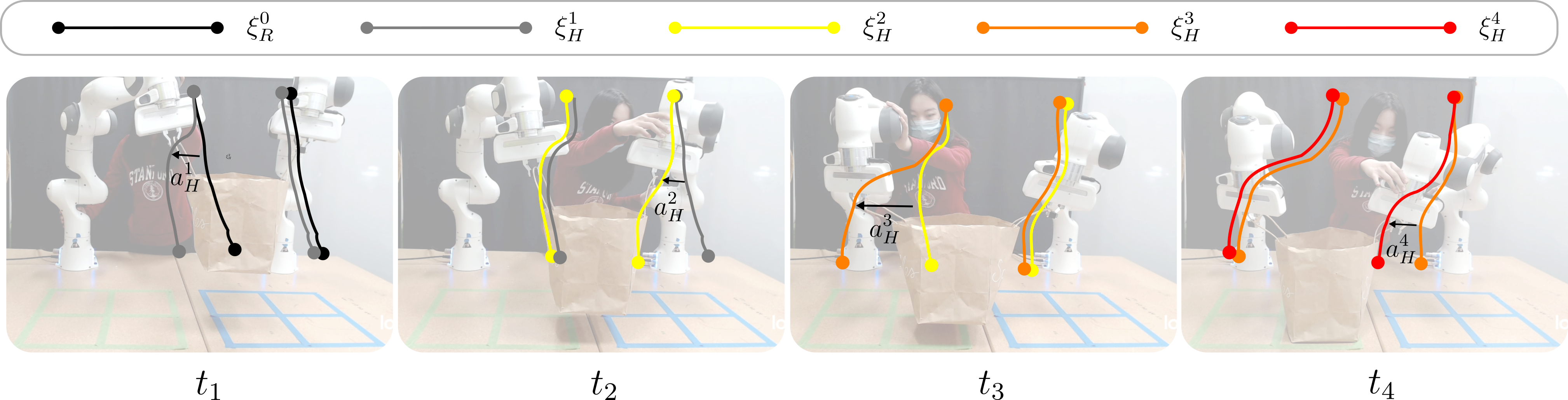}

		\caption{An example of a sequence of human corrections along with her corresponding correction trajectories $\xi_H^1, \xi_H^2, \xi_H^3, \xi_H^4$ to guide the robot to place the grocery bag on the green region while avoiding any stretching or squeezing of the bag.}
		\label{fig:propogation_traj}
	\end{center}
	
 	\vspace{-2em}
\end{figure*}

Leveraging these corrections, our goal is to find a better estimate of the reward parameters $P(\theta \mid A_H, \xi_R^0)$. We start by applying Bayes' rule:
\begin{equation} \label{eq:bayes}
   P(\theta \mid A_H, \xi_R^0) \propto P(\theta) P(A_H \mid \xi_R^0,\theta)
\end{equation}
In line with prior work \cite{bajcsy2017learning, bajcsy2018learning, losey2018including, bobu2020quantifying}, we will model $P(A_H|\xi_R^0, \theta)$ by mapping each human correction to a \textit{preferred trajectory}. Given the human's correction $(t_i, a_H^i)$, we deform the robot's trajectory to reach $\xi_H^i$. 
\mx{One simple example of this is to let the robot execute $a_R^{t_i} + a_H^{t_i}$ at this time step $t_i$, and stick to its original action plan $a_R^t$ for future time steps $t > t_i$.} More generally, we propagate the human's applied correction along the robot's current trajectory \cite{losey2017trajectory}:
\begin{eqnarray}
\centering
\begin{aligned}
    \xi_H^1 &= \xi_R^0 + \mu A^{-1} a_H^1\\
    \xi_H^i &= \xi_H^{i-1} + \mu A^{-1} a_H^i ,~~ i \in \{2,\ldots, K\}
\end{aligned}
\label{eq:traj_prop}
\end{eqnarray}
Consistent with \cite{losey2017trajectory} and \cite{dragan2015movement}, $\mu$ and $A$ are hyperparameters that determine the deformation shape and size. We emphasize that here the robot is not yet learning --- instead, it is locally modifying its trajectory in the direction of the applied correction. Within our motivating example, let the human apply a force pushing the first robot arm away from the blue region. Equation (\ref{eq:traj_prop}) maps this correction to $\xi_H$, a trajectory that moves the robot arm farther from the blue region than $\xi_R$. In Fig.~\ref{fig:propogation_traj}, we demonstrate how a sequence of corrections lead to a sequence of trajectories that enable the robot to correct its path and reach the preferred goals.


Now that we have this tool for mapping corrections to preferred trajectories, we can rewrite Equation~(\ref{eq:bayes}):
\begin{eqnarray}
\begin{aligned}
    P(\theta \mid A_H, \xi_R^0) &\propto P(\theta) P(A_H \mid \xi_R^0, \theta) \\
    &= P(\theta)P\Big((t_1, a_H^1), \ldots, (t_K, a_H^K) \mid \xi_R^0, \theta\Big) \\
    &\approx P(\theta)P(\xi_H^1, \ldots, \xi_H^K \mid \xi_R^0, \theta)
\end{aligned}
\label{eq:bayesian}
\end{eqnarray}
Here $P(\theta)$ is the robot's prior over the human's objective, and $P(\xi_H^1, \ldots, \xi_H^K \mid \xi_R^0, \theta)$ is the likelihood that the human provides a specific \textit{sequence} of preferred trajectories given the robot's initial behavior $\xi_R^0$ and the reward weights $\theta$.

\section{Learning from Sequences of Physical Corrections}
\label{sec:approach}

In the previous section we outlined how robots can learn from physical corrections using Equation~(\ref{eq:bayesian}). However, we still do not know how to evaluate $P(\xi_H^1, \ldots, \xi_H^K \mid \xi_R^0, \theta)$, which captures the relationship between a sequence of human corrections and the human's underlying objective. Prior work \cite{bajcsy2017learning, bajcsy2018learning, losey2018including, bobu2020quantifying} has avoided this problem by assuming that the human's corrections are \textit{conditionally independent}:
\begin{equation} \label{eq:independent}
    P(\xi_H^1, \ldots, \xi_H^K \mid \xi_R^0, \theta) = \prod_{t = 1}^K P(\xi_H^t \mid \xi_R^t, \theta)
\end{equation}
Intuitively, this assumption means that there is no relationship between the human's previous corrections and their current correction. But we know this is not always the case --- think about our motivating example, where the human's corrections are intricately coupled! Accordingly, here we propose a method for evaluating $P(\xi_H^1, \ldots, \xi_H^K \mid \xi_R^0, \theta)$ \textit{without} assuming conditional independence.

\subsection{Reasoning over Sequences of Physical Corrections}
\label{sec:approach:sequence}

To learn from sequences of human corrections online, we introduce an auxiliary reward function. In this section we also describe the modeling assumptions made by this auxiliary reward function, as well as an algorithm for leveraging this function for real-time inference.

\smallskip
\prg{Accumulated Evidence.} We start by introducing the auxiliary reward function: $D(\xi_H^1, \ldots, \xi_H^K, \theta)$. Let's refer to $D$ as the \textit{accumulated evidence} of a sequence of preferred trajectories $\xi_H^1, \ldots, \xi_H^K$ under reward parameter $\theta$. We hypothesize that the accumulated evidence should \textbf{(a)} reward not only the behavior of the final trajectory after all corrections, but also the intermediate trajectories the robot follows during the sequence of corrections. This recognizes that --- when users make corrections --- they don't sacrifice long-term reward for short-term failure. Consider our motivating example: the human is not willing to correct the robot into crushing the bag, even if that will reduce the overall number of corrections needed to avoid the obstacle. Of course, \textbf{(b)} humans also try to minimize their overall effort when making corrections --- and any rewards for which the human's corrections are redundant or unnecessary are therefore not likely to be the human's true reward. Combining these two terms:
\begin{multline}
D\big(\xi_H^1, \ldots, \xi_H^{k},  \theta\big) =\\   
\sum_{t = 1}^{K} \alpha ^{K-t}R(\xi_H^t, \theta) - \gamma \Big(\sum_{t = 1}^K\|a_H^t\|^2\Big) 
\label{eq:D_def}
\end{multline}
$\alpha$ and $\gamma$ are two hyperparameters decaying the importance of previous corrections, and determining the relative trade-off between intermediate reward and human effort respectively.

\smallskip
\prg{Learning Rule.} Similar to prior work in inverse reinforcement learning \cite{ziebart2008maximum, ramachandran2007bayesian, osa2018algorithmic, jeon2020reward}, we model humans as noisily rational agents whose corrections maximize accumulated evidence for their preferred $\theta$:
\begin{equation} \label{eq:likely}
    P(\xi_H^1, \ldots, \xi_H^{K} \mid \xi_R^0, \theta) \propto \exp\Big(D\big(\xi_H^1, \ldots, \xi_H^{k}, \theta\big)\Big)
\end{equation}
Equation (\ref{eq:likely}) expresses our likelihood function for learning from human corrections. Using Laplace's method (as applied in \cite{dragan2013policy}), we can approximate the normalization factor:
\begin{flalign}
\begin{aligned}
&P(\xi_H^1, \ldots, \xi_H^{K} ~|~ \theta)\\
& =\frac{\exp\Big(D\big(\xi_H^1, \ldots, \xi_H^{k}, \theta\big)\Big)}{\int_{\xi_H^1, \ldots, \xi_H^{K}}\exp\Big(D\big(\xi_H^1, \ldots, \xi_H^{k}, \theta\big)\Big)d\xi_H^1 \ldots d\xi_H^{K}} \\
& \approx \frac{\exp\Big(D\big(\xi_H^1, \ldots, \xi_H^{k}, \theta\big)\Big)}{\exp\Big(\max_{ \xi_H^1, \ldots, \xi_H^{K}}D\big(\xi_H^1, \ldots, \xi_H^{k}, \theta\big)\Big)}.
\end{aligned} &&
\label{eq:full_normalizer}
\end{flalign}  

\smallskip
\prg{Monte-Carlo Mixed-Integer Optimization.} Inspecting the denominator of Equation (\ref{eq:full_normalizer}), we see that --- in order to evaluate the likelihood of a sequence of corrections --- we need to find the \textit{highest possible} accumulated evidence the human \textit{could} have achieved given that their objective is $\theta$. Put another way, we need to search for the sequence of $K$ corrections that maximize Equation~(\ref{eq:D_def}) under $\theta$:
\begin{eqnarray}
\begin{aligned}
    D_K^*(\theta) &= \max_{ \xi_H^1, \ldots, \xi_H^{K}}D\big(\xi_H^1, \ldots, \xi_H^{k}, \theta\big) \\
    &= \max_{ (t_1, a_H^1), \ldots, (t_K, a_H^{K})}D\big(\xi_H^1, \ldots, \xi_H^{k}, \theta\big)
    \label{eq:optim_objective}
\label{eq:D_optimization}
\end{aligned}
\end{eqnarray}
Unfortunately, solving Equation~(\ref{eq:optim_objective}) is hard because we need to figure out both \textit{when} to make each correction ($t_1, \ldots, t_K$), which is a discrete decision, as well as \textit{what} to correct during each interaction ($a_H^1, \ldots, a_H^K$), which is a continuous action. 

To tackle this mixed-integer optimization problem, we develop a Monte-Carlo mixed-integer optimization method, where we first randomly sample discrete ($t_1, \ldots, t_K$), and then solve ($a_H^1, \ldots, a_H^K$) with gradient-based continuous optimization methods. The full pipeline for the optimization is summarized in Algorithm~\ref{algo:mixed_optimization}.

Importantly, the inputs to this optimization are only the robot's initial trajectory $\xi_R^0$, the reward parameter $\theta$, and our model hyperparameters. Hence, we can conduct \textit{offline} optimization to solve Equation~(\ref{eq:D_optimization}) for several sampled values of $\theta$. We then use the resulting library of stored $D^*$ values to learn online, during physical interaction.

\begin{algorithm}[t]
 \caption{Monte-Carlo Mixed-Integer Program
 }
 \label{algo:mixed_optimization}
 \SetAlgoLined
 \KwOut{Maximum accumulated evidence $D_K^*(\theta)$, and $K$ optimized human corrections $(t_1, a_H^1), \ldots, (t_K, a_H^{K})$}
 \KwIn{The suboptimal initial robot plan $\xi_R^0$, reward parameter $\theta$, and hyperparmeters in \eref{eq:D_def}. \newline $G(t_1, \ldots, t_K)$ is a continuous optimizer.}
 Initialize maximum accumulated evidence and correction times: $D_K^* = -\infty$, $T = \emptyset$.\\
 
 \For{$i\gets0$ \KwTo $T_{\text{max}}$}{
    \While{$(t_1,\ldots, t_K) \in T$}{
        Randomly sample $(t_1,\ldots, t_K)$
    } 
    $T = T \cup \{(t_1,\ldots, t_K)\}$
    
    $D_i ~,~ (a_1, \ldots, a_K) = G(t_1, \ldots, t_K, \xi_R^0, \theta)$
    
    \If{$D_i > D_K^*$}{
        $D_K^* = D_i$

        $T_H^* = (t_1, \ldots, t_K)$
        
        $A_H^* = (a_1, \ldots, a_K)$
        }
}
return $D_K^*, T_H^*, A_H^*$
\end{algorithm}


 
 
    
    

        

\smallskip
\prg{Online Inference.}
We have a way for solving for the denominator in Equation~(\ref{eq:full_normalizer}) offline. What remains to be evaluated is the numerator $\exp{\Big(D\big(\xi_H^1, \ldots, \xi_H^{k}, \theta\big)}\Big)$. This can be easily computed online when the human is physically correcting the robot. Thus, we can now evaluate $P(\xi_H^1, \ldots, \xi_H^{K} \mid \xi_R^0, \theta)$ while accounting for the relationships between corrections.

Our overall pipeline is as follows. Offline, we compute $D_K^*(\theta)$ with Alg.~\ref{algo:mixed_optimization}.
Online, the robot starts by following the optimal trajectory $\xi_R^0$ \mx{with respect to its prior over $\theta$. However, this initial reward might not capture the human reward and thus the human physically corrects the robot.} \mx{The robot would then perform inference and update its belief over the reward when it receives new human corrections.}
At time $t$, the human has provided a total of $K$ corrections $(t_1, a_H^1), \ldots, (t_{K}, a_H^{K})$, where $t_1 < t_2 < \cdots <t_{K} \leq t$. We propagate these human corrections to get the deformed trajectories $\xi_H^1, \ldots, \xi_H^{K}$ with Equation~(\ref{eq:traj_prop}).
We then perform Bayesian inference by solving Equation~(\ref{eq:full_normalizer}) and plugging the likelihood function into Equation~(\ref{eq:bayesian}).
Finally, the robot uses its new understanding of $\theta$ to solve for an updated optimal trajectory $\xi_R^t$.



\subsection{Relation to Prior Works: Independent \& Final Baselines}
To evaluate the effectiveness of our proposed method that reasons over correction sequences, we compare against two baselines: \emph{Independent} and \emph{Final}.

\begin{figure*}[th!]
	\begin{center}
		\includegraphics[width=2.0\columnwidth]{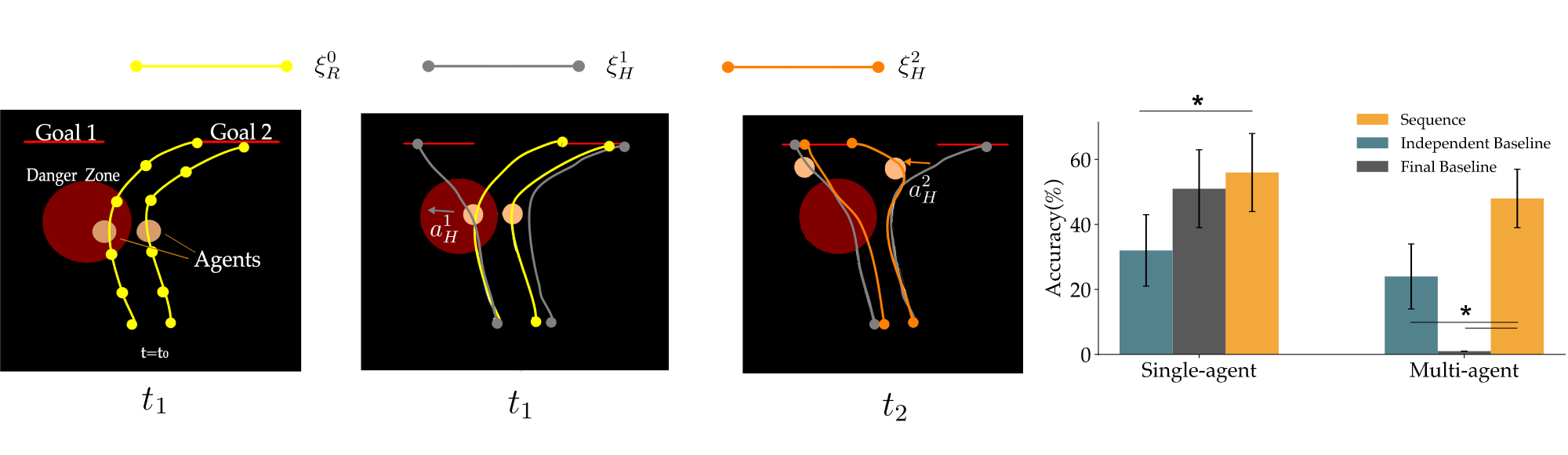}
\vspace{-1em}
		\caption{Navigation Simulation. We show the sequence of corrections in the multi-agent task, and the accuracy results of the single- and multi-agent tasks.}

		\label{fig:results_navi}
	\end{center}
	
	\vspace{-2em}
\end{figure*}

\smallskip
\prg{Independent Baseline (Online).}
This baseline follows the same formalism as our approach, but assumes each human correction is conditionally independent. Hence, here we use Equation~(\ref{eq:independent}) to learn from each correction separately. The likelihood of observing an individual correction is related to the reward and effort associated with that correction \cite{bajcsy2017learning}:
\begin{equation}
    \label{eq:ind}
    P(\xi_H^i \mid \xi_R^{i}, \theta) \propto \exp\big(R(\xi_H^i, \theta) - \gamma\|\xi_H^i - \xi_R^i\|^2\big)
\end{equation}

\smallskip
\prg{Final Baseline (Offline).}
At the other end of the spectrum, we can always look at the final trajectory when all corrections are finished \cite{jain2015learning}. In this case, the robot learns by comparing the final trajectory, $\xi_H^K$, to the initial trajectory, $\xi_R^0$:
\begin{equation} \label{eq:final}
    P(\xi_H^K \mid \xi_R^0, \theta) \propto \exp\big(R(\xi_H^K, \theta) - \gamma\|\xi_H^K - \xi_R^0\|^2\big)
\end{equation}
To summarize, both our method and these baselines use a Bayesian inference approach. The difference is the likelihood function: \textit{Independent} assumes conditional independence, while \textit{Final} only considers the initial and final trajectory.

\section{Experiments}
\label{sec:user_study}

To test our proposed algorithm, we conduct experiments with human users in a simulated navigation task and a robot manipulation task. We will discuss details that are consistent in both tasks and then elaborate on each one respectively.

\smallskip
\prg{Tasks.} Our two tasks are: a web-based online simulated navigation task and an in-person robot manipulation task.


\noindent\textit{1) Navigation Simulation}. 
In this task, a team of robots are navigating together through a specified region as in \figref{fig:results_navi}. The robot’s objective function considered four features related to: reaching the goal, keeping formation, avoiding the danger zone, and minimum travel distance. We test our algorithm and baselines in different scenarios by varying robot team size, robot initial policy, and specifying different human preference reward values. 
\\
\noindent\textit{2) Robot Manipulation}. In this task, two robot arms are carrying a full grocery bag to place it on the table. There are four features concerned in the reward: reaching the goal basket (blue or green regions as shown in~\figref{fig:front}), keeping the groceries inside of the bag while avoiding squeezing or stretching the bag, avoid touching nearby housewares such as cabinets or the hat shown in~\figref{fig:front}, and minimum trajectory length for efficiency.
The robot starts off with the assumption of reaching an incorrect goal while also not realizing the bag is full, and should not be stretched or squeezed. Users apply forces to guide the two robot arms toward the correct goal region, while trying to keep the groceries inside of the bag. 

\begin{figure*}[th!]
	\begin{center}

\includegraphics[width=2.0\columnwidth]{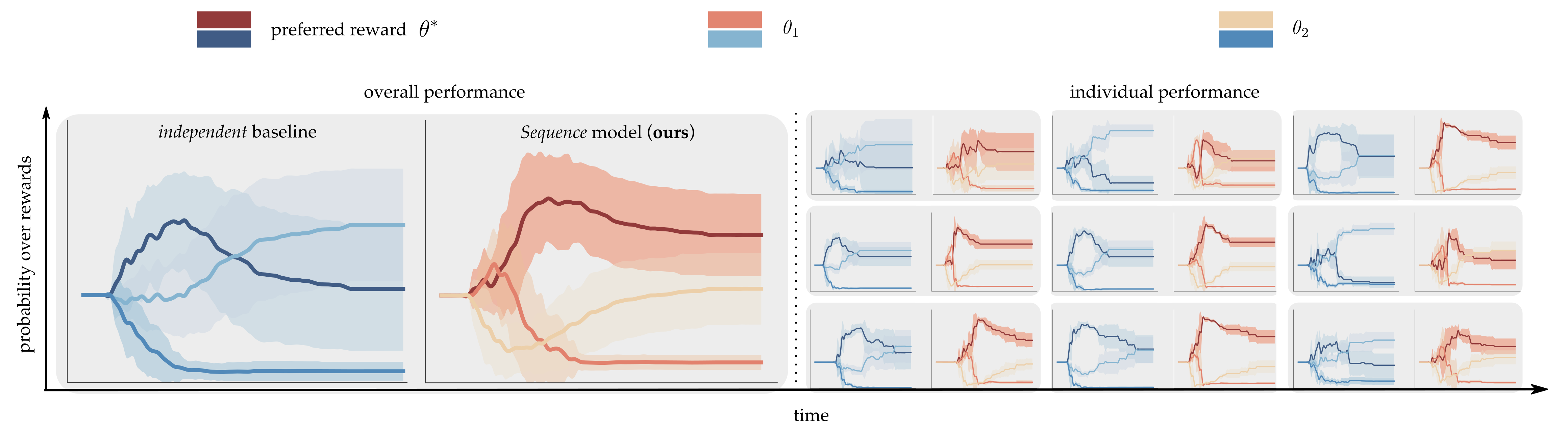}

		\caption{Likelihood of three different reward parameters ($\theta^*, \theta_1, \theta_2$) as more corrections are received over time. The preferred reward $\theta^*$ is shown with the darker red or blue.  Each pair of plots demonstrate the \emph{Sequence} model compared with the \emph{Independent} model. We demonstrate the aggregate results over all users on the left, and the individual performance on the right. As shown \emph{Sequence} outperforms \emph{Independent} in identifying the preferred reward $\theta^*$.}

		\label{fig:align}
	\end{center}
	
	\vspace{-2em}
\end{figure*}

\prg{Independent Variables.} We compared three different inference models: reasoning over corrections independently (\textit{Independent}), performing inference only based on the final correction (\textit{Final}), and our model that reasons over the sequence of corrections (\textit{Sequence}). \textit{Independent} and our \textit{Sequence} model can perform online inference, while \textit{Final} will conduct offline inference after all corrections are provided.


\smallskip
\prg{Dependent Measures.}
We conduct experiments with human users and evaluate the effectiveness of the models by measuring the inference accuracy. Since we are unable to measure users' internal reward, we specify users' preferred reward function out of a predefined finite set of candidate reward parameters. We convey the preferred reward by explaining the priorities of the features to the user and demonstrating a desired robot trajectory using the preferred reward.
Users are instructed to correct the robot to behave as optimally as possible, while minimizing their physical correction effort. 
\smallskip

\prg{Hypotheses}: 
\begin{hypothesis}
    Compared to the online \emph{Independent} baseline, reasoning over \emph{Sequences} of corrections leads to higher accuracy and faster convergence to the preferred reward.
\end{hypothesis}
\begin{hypothesis}
    Compared to the offline \emph{Final} baseline, in addition to the advantages of online inference, our \emph{Sequence} model achieves higher accuracy in challenging tasks -- specifically tasks where fully correcting the robots is infeasible.
\end{hypothesis}
\smallskip



	



	

\vspace{-1em}
\subsection{Navigation Simulation}
\label{sec:exp_navigation}
\prg{Experimental Details.}
We recruited $15$ participants for two simulated navigation tasks. Participants interact with point-mass robots using a web browser, where they can observe the robots' trajectories, select a robot to correct, and provide corrections using the arrow keys.
We collected data from humans for $5$ episodes in two different scenarios: a simple single-agent scenario with only one robot, and  a more complex scenario containing a two-robot team. 
In both settings, participants are only able to correct one robot at a time.

In both scenarios, robots' initial trajectory goes to an incorrect goal region as shown in Fig.~\ref{fig:results_navi}.
In the human's preferred reward function, not only is going to the correct goal encouraged, but other features are also encoded, including avoiding a danger zone and keeping the formation (equal distance between the robots throughout the trajectories). 

\prg{Results \& Analysis.}
We compute the average inference accuracy for the two baseline methods and our method across all users in both scenarios. The results are shown in \figref{fig:results_navi}.

Across both scenarios, our \textit{Sequence} model demonstrates leading performance compared to both the \textit{Independent} and the \textit{Final} baselines. One thing to note is that in the single-agent scenario, the \textit{Final} method has a comparable accuracy to our method.
This indicates that in this simple task, the user can correct the robot to behave almost optimally so that simply looking at the final trajectory conveys sufficient information about the preferred reward. While in a more complex multi-agent scenario, the performance of the \textit{Final} method drops significantly
. 
This is because in this complex scenario, even if the user acts optimally, the final trajectory will still not have sufficient information to identify the preferred reward. In this example reasoning over sequences is dominant over only considering the final correction.

\subsection{Robot Manipulation}
\label{sec:exp_robot}
\prg{Experimental Details.}
We recruited $9$ participants for the in-person robot manipulation task (Fig.~\ref{fig:front}). Here, the two Franka Emika Panda robot arms are carrying a grocery bag to the table. 
The participants are asked to physically push or pull the robot to correct its behavior. Similar to the navigation task, the user can only correct one robot at a time, and we collected $5$ episodes of corrections from each participant. 

As shown in \figref{fig:front}, there are two containers on the table for the grocery bag (the blue region and the green region). The robots' initial plan is to carry the bag to the right toward the blue region. However, the human wants to put the grocery to the left container. Meanwhile, since the grocery bag is almost full, in order to keep the groceries from falling out, the participants are also instructed not to squeeze or stretch the bag. In this setting there are three possible reward parameter $\theta$, and the robot tries to infer the correct reward parameter $\theta^*$ from the human corrections. 

\smallskip
\prg{Results \& Analysis.}
We calculate the inference accuracy for all the three models (\emph{Independent}, \emph{Final}, and \emph{Sequence}), and summarize the results in \tabref{tab:robot_accuracy}. Our method demonstrates superior performance compared to the baselines.

\begin{table}[!h]
\centering
\caption{Inference accuracy over 9 participants for robot manipulation.}
\begin{tabular}{@{}cccccl@{}}
\hline
\noalign{\vskip 0.5mm}
\noalign{\vskip 0.5mm}
       & \textit{Sequence} (\textbf{ours})         & \textit{Independent}           & \textit{Final}   \\ 
       \hline\noalign{\vskip 1mm}  
accuracy (\%)       & $\textbf{82.22}\pm \textbf{21.99}$        & $31.11\pm26.99$          & $53.33\pm13.30 $      \\ \noalign{\vskip 1mm}  \hline\noalign{\vskip 1mm}  

\end{tabular}
\label{tab:robot_accuracy}
\end{table}

In addition, we illustrate the probability distribution over $\theta$ with time in \figref{fig:align}. Since the \emph{Final} baseline performs the inference offline, we only visualize our \textit{Sequence} model and the \textit{Independent} baseline. As can be seen, across all of the participants, the probability for the preferred reward consistently dominates the other candidate rewards. However, for the \textit{Independent} baseline, even if the probability for the preferred reward sometimes starts off high, it ends up not being the most likely reward as we receive more corrections.
This is because in this two-robot task, redirecting the system to the correct goal while simultaneously maintaining the shape of the bag (formation) is not possible. The best possible corrections are: push one arm towards the goal, while stretching or squeezing the bag by a small amount, and then push the other arm in the same direction so that the bag shape is recovered. However, if we reason over these corrections independently, the robot will think that the first push indicates that preserving the bag shape is not important, and this leads to an incorrect inference.

\subsection{Summary} 
Our results empirically support both of our hypotheses \textbf{H1} and \textbf{H2}. Our \textit{Sequence} model not only conducts an online inference, but also demonstrates superior performance specially in complex multi-agent tasks. 
\section{Conclusion}
\label{discussion}

We developed a framework for learning from sequences of user corrections during physical human-robot interaction. We introduced an auxiliary reward that models the connections between corrections, and leveraged mixed-integer programming to solve for the best possible sequence of corrections. Our results from online and in-person users demonstrate that our approach outperforms methods that take each correction independently, or wait for the final trajectory.







\clearpage
\section*{Acknowledgement}
We acknowledge funding from the NSF Award \#1941722 and \#2006388, Future of Life Institute, and DARPA Hicon-Learn project.

\balance
\bibliographystyle{IEEEtran}
\bibliography{main}

\end{document}